\pdfoutput=1

\documentclass[11pt]{article}

\usepackage[final]{acl}

\usepackage{times}
\usepackage{latexsym}

\usepackage[T1]{fontenc}

\usepackage[utf8]{inputenc}

\usepackage{microtype}

\usepackage{inconsolata}

\usepackage[ruled]{algorithm2e}
\usepackage{graphicx}
\usepackage{multirow}
\usepackage{latexsym}
\usepackage{amsmath,amssymb,amsthm}
\usepackage{subfigure}
\usepackage{microtype}
\DeclareGraphicsExtensions{.pdf, .png}
\usepackage{epstopdf}

\usepackage{booktabs}
\usepackage{setspace}
\usepackage[normalem]{ulem}
\useunder{\uline}{\ul}{}
\usepackage{tikz}
\usepackage{bbding}
\usepackage{pifont}
\usepackage{wasysym}
\usepackage{paralist}
\usepackage{setspace}

\title{Retrieving Implicit and Explicit Emotional Events Using Large Language Models}
\author{Guimin Hu$^{\spadesuit}$, Hasti Seifi$^{\dagger}$\\
  $^{\spadesuit}$University of Copenhagen,\\
  $^{\ddagger}$Arizona State University, United States\\
  \texttt{hasti.seifi@asu.edu, rice.hu.x@gmail.com}}
\begin{document}
\maketitle

\begin{abstract}
Large language models (LLMs) have garnered significant attention in recent years due to their impressive performance. While considerable research has evaluated these models from various perspectives, the extent to which LLMs can perform implicit and explicit emotion retrieval remains largely unexplored. To address this gap, this study investigates LLMs’ emotion retrieval capabilities in commonsense. Through extensive experiments involving multiple models, we systematically evaluate the ability of LLMs on emotion retrieval. We propose a supervised contrastive probing method to evaluate large language models (LLMs) in retrieving implicit and explicit emotional events for specific emotions, as well as assessing the diversity of the retrieved emotional events. Our findings provide valuable insights into the strengths and limitations of LLMs in emotion-focused retrieval tasks.
\end{abstract}


\section{Introduction}
Large language models (LLM) have drawn considerable interests in recent years due
to their remarkable performance. In the constantly evolving realm of artificial intelligence, Large Language Models (LLMs) have risen as cutting-edge tools for natural language understanding. In community of sentiment analysis, LLMs’ emotion retrieval abilities come under scrutiny of the commonsense retrieval domain. Past research has delved into the potential of language models for emotion recognition~\cite{hu2024unimeec}, emotion cause analysis~\cite{zheng2022ueca,DBLP:journals/kbs/HuLZ21} and sentiment analysis~\cite{DBLP:conf/emnlp/HuLZLWL22}, as seen in studies~\cite{DBLP:journals/corr/abs-2108-11626,DBLP:conf/icassp/HuHWJM22,DBLP:conf/acl/XiaD19}. 

\begin{table}[t]
\centering
\resizebox{\linewidth}{!}{
\begin{tabular}{llcc}
\toprule
{\bf Emotion} & {\bf Emotional Event} & {\bf Explicit} & {\bf Implicit}              \\
\midrule
Joy     & be proud of someone & \XSolidBrush & \Checkmark \\
Joy     & successful career& \XSolidBrush & \Checkmark \\
Joy     & glad someone helped& \Checkmark  & \XSolidBrush\\
Sad     & feel lonely& \XSolidBrush & \Checkmark \\
Sad     & I feel like a loser& \XSolidBrush & \Checkmark \\
Sad     & sad because they didn't win& \Checkmark  & \XSolidBrush\\
Angry     & angry at not being able to control oneself& \Checkmark &\XSolidBrush\\
Angry     & feel angry at someone's laziness& \Checkmark  & \XSolidBrush\\
Angry    & he lost his temper& \XSolidBrush & \Checkmark \\
\bottomrule
\end{tabular}}
\caption{Examples of Emotional Events.}
\label{tab:some_cases}
\end{table}

Generally, receiving praise typically elicits a feeling of happiness in the individual. On the contrary, criticism often leads to feelings of pessimism and frustration. We refer to this type of event, which can trigger a specific emotion in humans, as an ``\textbf{emotional event}''. From this perspective, the reasoning linking emotional events and triggered emotions overlaps with common-sense reasoning in certain situations (for example, it is common sense that most people feel happy when praised).Previous research on sentiment analysis, emotion category detection, and emotion cause extraction has demonstrated the capacity of LLMs in emotion-related tasks. However, these studies typically use LLMs as a backbone, training models with supervised learning and evaluating their performance on tasks like recognizing emotions in text or extracting emotion causes, while overlooking the exploration of LLMs' ability to perform emotion retrieval within the realm of common sense. To address this gap, our study provides both qualitative and quantitative analyses of LLMs' ability to reason implicit and explicit emotional events in common-sense contexts. Implicit emotional events refer to situations where no explicit sentiment words are present regarding the topic.
Our goal is to evaluate model performance in retrieving emotional events across various emotion categories, both implicit and explicit, with a focus on answering the following three questions:

\paragraph{RQ1: Assessment of LLMs on emotional events retrieval across different emotion categories} 
To address this question, we evaluate the capacity of LLMs in emotional event retrieval by using specific emotions (e.g., ``Joy'', ``Sad'', and ``Angry'') as input to retrieve emotional events relevant to the given emotion.
\paragraph{RQ2: What are the differences in retrieval performance for implicit and explicit emotional events across different LLMs?} To answer this question, we evaluate the accuracy with which LLMs retrieve implicit and explicit emotional events. 
\paragraph{RQ3: How does the diversity of retrieved emotional events vary across LLMs at different levels of precision?}
To address this, we employ the supervised contrastive probe to investigate the retrieval ability of LLM and conduct experiments to evaluate the abilities of LLM's retrieving diverse emotional events.

In summary, our contributions to tackle these challenges are:
(1) we conduct a comprehensive evaluation of state-of-the-art Large Language Models (LLMs) in emotional event retrieval, with a specific focus on implicit and explicit emotion retrieval, using public commonsense knowledge base, (2) our analysis is thorough and multifaceted, encompassing both qualitative and quantitative aspects into the strengths and limitations of LLMs in emotion retrieval of commonsense, and (3) we introduce a novel supervised contrastive probe and this probe is tested on its ability to generalize to objects that were not seen during training.

\section{Dataset}
We use the emotion-cause flow in $\text{C}^{3}\text{KG}$~\cite{li2022c3kg}, a Chinese commonsense knowledge base that includes concept flow, event flow, emotion intention flow, and emotion-cause flow.
We extract emotional events corresponding to specific emotions from $\text{C}^{3}\text{KG}$ based on the emotion-cause flow. In the emotion-cause flow, each emotion category is associated with multiple emotional events. Additionally, we manually annotate the emotional events as explicit or implicit. Table \ref{tab:distribution_c3kg} shows the distribution of emotion categories across the training, validation, test sets and the distribution of explicit emotional events for each category.
\begin{table}[t]
\centering
\resizebox{\linewidth}{!}{
\begin{tabular}{lccccc}
\toprule
 & Train & Valid & Test  & Total & Explicit         \\
\midrule
joy     &1922  & 247 & 540 & 2704 & 1305\\
sad     &370 & 64 & 256 & 1245 & 171 \\
angry     &851 & 138  & 102 &541 & 173\\
total &3143&449 &898 &4481 & 1649\\
\bottomrule
\end{tabular}}
\caption{Distribution of emotion categories across train, valid, and test datasets.}
\label{tab:distribution_c3kg}
\end{table}

\section{Method}
\subsection{Problem Formalization}
Given an emotion category, the emotional event retrieval task aims to retrieve events associated with the specified emotion. Our objective is to assess this ability across LLMs from the following perspectives: a) the precision of retrieved emotional events for a specific emotion, b) the precision of retrieved emotional events for implicit and explicit emotions, and c) the diversity of retrieved emotional events.
\subsection{A Supervised Contrastive Probe}
\begin{figure*}[t]
\centering
\subfigure[\label{fig:visualization:a}]{
\includegraphics[width=\textwidth]{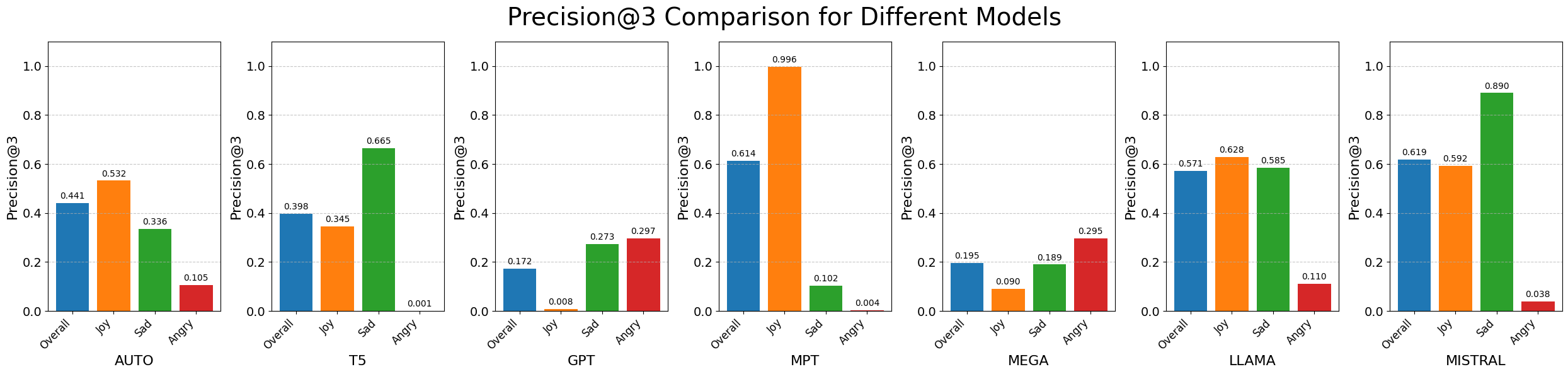}}
\hspace{0.05in}
\subfigure[\label{fig:visualization:b}]{
\includegraphics[width=\textwidth]{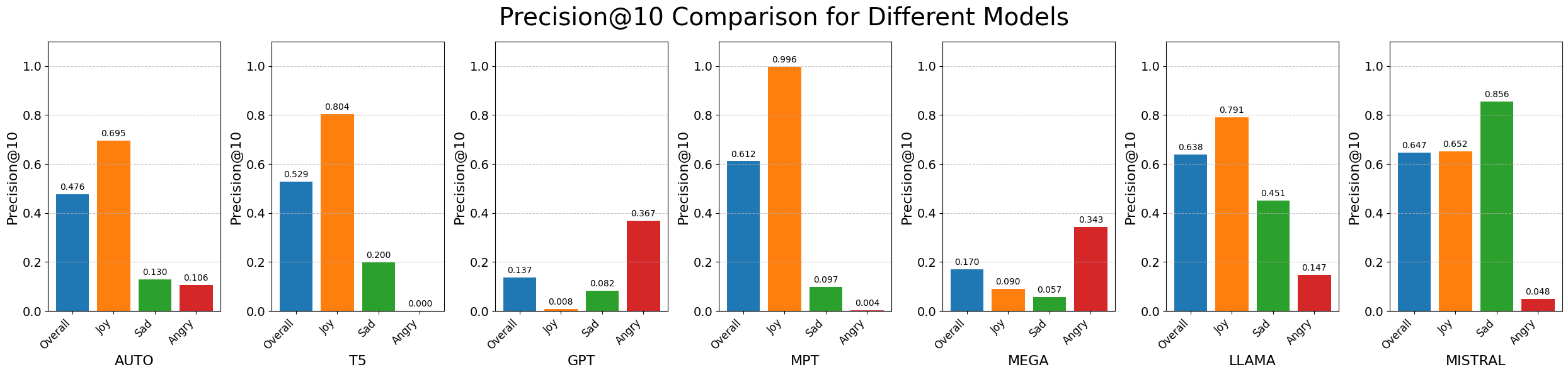}}
\hspace{0.05in}
\subfigure[\label{fig:visualization:c}]{
\includegraphics[width=\textwidth]{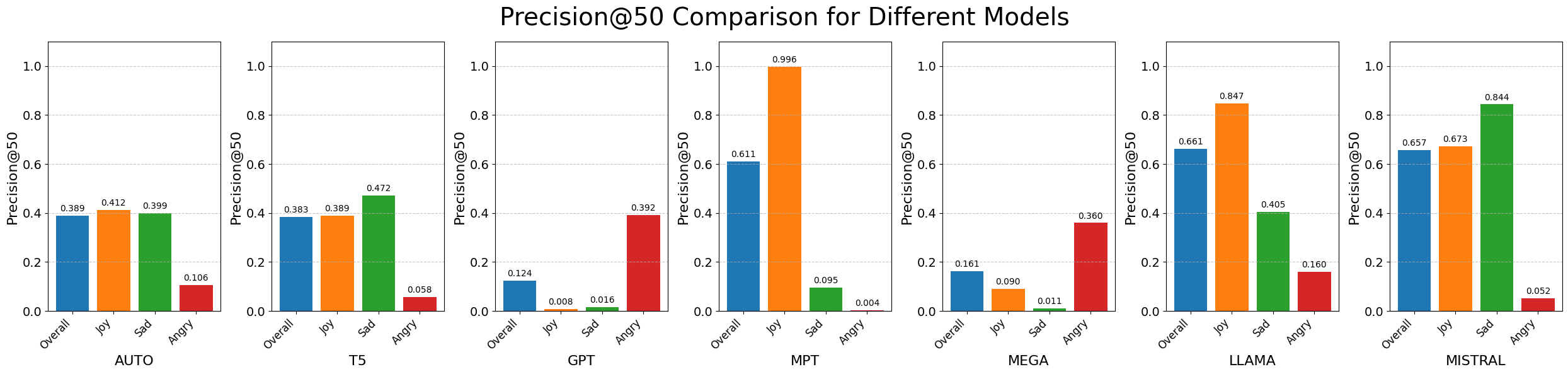}}
\caption{Performance (P@3, P@10, P@50) of different LLMs on the emotion retrieval.}
\label{fig:main_results}
\end{figure*}

Each emotional event corresponds to a specific emotion category (e.g., ``successful career brings someone joy''). We treat the emotion category as the label for the emotional event. 

Given an anchor sample with emotion category $e_{i}$, emotional event $c_{j}\in \mathcal{C}$ sharing the same emotion category form positive contrastive pairs. In supervised contrastive learning, the anchor is associated with multiple positive and negative samples. Positive samples are emotional events labeled with the same emotion category as the anchor, while negative samples are those labeled with a different emotion category. During training phase, we use supervised contrastive learning loss as the loss objective to train a supervised contrastive probe.

Suppose we partition dataset $\mathcal{D}$ into mutually exclusive train and test sets, $\mathcal{D}=\mathcal{D}_{train}\cup \mathcal{D}_{valid}\cup \mathcal{D}_{test}$. If we learn $\{W_{1}, W_{2}\}$ via minimizing training loss and these transformations generalize to $\mathcal{D}_{test}$, i.e., for $c\in \mathcal{D}_{test}$:
\begin{align}
\begin{split}
    &c = \mathop{\mathcal{ARGMAX}} \text{sim}(e_{i}, c_{j}), e_{i}\in \mathcal{D}_{test}\\
 &\text{sim}(e_{i}, c_{j})=\frac{\left \langle W_1f_{LM}(e_{i}), W_2f_{LM}(c_{j}) \right \rangle}{\parallel W_1f_{LM}(e_{i})\parallel \parallel W_2f_{LM}(c_{j})\parallel}\\
\end{split}
\end{align}
To evaluate the precision of retrieved emotional events for a given emotion, we use precision@K over $\mathcal{D}_{test}$. A prediction is considered correct if the correct label is among the top K most similar objects in the retrieval set.

\section{Experiments}
\subsection{Setup}
\paragraph{LLMs} We adopt LLMs including BERT~\cite{kenton2019bert}, T5~\cite{DBLP:journals/jmlr/RaffelSRLNMZLL20}, GPT~\cite{brown2020language}, MPT~\cite{muennighoff2024olmoe}, Mega, Llama~\cite{touvron2023llama}, Mistral\footnote{https://huggingface.co/mistralai/Mistral-7B-Instruct-v0.3} to examine emotional event retrieval performance.
\paragraph{Precision and Diversity}
We adopt Precision@K and Diversity@K (D@K) to evaluate the model performance, where Precision@K (P@K) counted as correct of the correct label is in the set of highest K scores and Diversity@K (D@K) represents the diversity of emotional events correctly retrieved among the top K results. The metrics are given as $P@K = \frac{n_{cr}^{K}}{n_{ar}}, D@K = \frac{n_{ur}^{K}}{n_{cr}^{K}}$, where $n_{cr}^{K}$ denotes the number of relevant emotional events in the set of highest K scores, $n_{ar}$ denotes the number of all retrieval results, and $n_{ur}^{K}$ denotes the number of de-duplicated emotional events in the set of highest K scores.
    

\subsection{Main Results}
Figure \ref{fig:main_results} shows the performance comparison of different Large Language Models (LLMs). ``Joy'' consistently has the highest precision across all models, with MPT and MISTRAL being the top performers. Precision drops sharply for emotions like ``Sad'' and ``Angry'' across all models, indicating that emotion retrieval for these emotions is more challenging. MPT consistently shows the highest performance for ``Joy'' in all settings, while MISTRAL also performs well, though slightly behind MPT. The results suggest that some models like MPT and MISTRAL perform better for detecting ``Joy'' but struggle with other emotions, particularly ``Angry''. Models like GPT show consistently poor performance across the board, indicating weaker ``Angry'' events retrieval capabilities.



\subsection{Implicit VS Explicit Emotion Retrieval}
Table \ref{tab:explicit_results} presents the diversity of retrieval results for various models (BERT, T5, GPT, etc.) across three emotional categories: Joy, Sadness, and Anger at different precision levels (P@3, P@10, P@50). As K increases, most models show a noticeable increase in diversity, especially for ``Joy'' and ``Sadness''. Mistral stands out for achieving the highest diversity across all emotional categories, especially in the retrieval of "Sadness". Retrieval for ``Joy'' is generally the highest across models, whereas ``Sadness'' and ``Anger'' have lower retrieval rates, especially at lower K values. The retrieval on ``Sadness'' is particularly weak across all models, with very few models succeeding in retrieving it.

\begin{table}[]
\centering
\resizebox{0.85\linewidth}{!}{
\begin{tabular}{llccc}
 \toprule
Model& P@K & Joy(\%)     & Sad(\%)    & Angry(\%) \\
\midrule
\multirow{3}{*}{BERT}&
P@3& 50.23& 0.0      & 0.0 \\
&P@10& 76.93 & 0.0      & 39.78 \\
&P@50& 53.10 & 20.33      & 29.23 \\
\midrule
\multirow{3}{*}{T5}
&P@3& 96.14 & 0.0      & 0.0      \\
&P@10& 75.11 & 0.0      & 0.0     \\
&P@50& 38.93 & 47.15      & 5.78      \\
\midrule
\multirow{3}{*}{GPT}
&P@3 & 66.67   & 0.0      & 35.05 \\
&P@10 & 70   & 0.0     & 35.05 \\
&P@50 & 70   & 0.0      & 35.05 \\
\midrule
\multirow{3}{*}{MPT}
&P@3 & 66.67   & 0.0      & 35.05 \\
&P@10 & 70   & 0.0     & 35.05 \\
&P@50 & 48.51   & 13.69     & 30 \\
\midrule
\multirow{3}{*}{Mega}
&P@3 & 66.67    & 0.0      & 36.42 \\
&P@10 & 70    & 0.0      & 36.42 \\
&P@50& 50    & 0.0      & 29.50 \\
\midrule
\multirow{3}{*}{Llama}
&P@3& 30.53 & 0.0 & 0.0 \\
&P@10& 76.18 & 8.73 & 33.5  \\
&P@50& 48.80 & 0.0 & 28.38  \\
\midrule
\multirow{3}{*}{Mistral}
&P@3  & 54.35 & 0.0 & 0.0\\
&P@10  & 72.78 & 9.82 & 34.37\\
&P@50  & 48.40 & 13.62 & 27.61\\
\bottomrule
\end{tabular}}
\caption{Retrieval rate of explicit emotion.}
\label{tab:explicit_results}
\end{table}

\begin{table}[]
\centering
\resizebox{\linewidth}{!}{
\begin{tabular}{llccc}
\toprule
Model &D@K& Joy(\%)     & Sad(\%)    & Angry(\%) \\
\midrule
\multirow{3}{*}{BERT}
 & D@3& 0.56 (3) & 0.8 (2) & 3.5 (4) \\
  & D@10& 1.9 (10) & 0.8 (2) & 9.8 (11)\\
   & D@50& 13.5 (72) & 17.39 (44)& 80.35 (90) \\
\midrule
\multirow{3}{*}{T5}
  &D@3& 0.56 (3) & 0.8 (2) & 0.89 (1)     \\
  &D@10& 1.9 (10) & 0.8 (2) & 0.0 (0)  \\
  &D@50& 7.31 (39) & 14.22 (36) & 16.96 (19) \\ 
\midrule
\multirow{3}{*}{GPT}
  &D@3& 0.56 (3) & 1.58 (4) & 2.6 (3) \\
  &D@10& 1.9 (10) & 0.4 (1) & 9.8 (11) \\
  &D@50& 9.3 (50) & 0.4 (1) & 67.85 (76) \\
\midrule
\multirow{3}{*}{MPT}
  &D@3& 0.75 (4)  & 1.5 (4)  & 2.6 (3)\\
  &D@10& 2.1 (11)  & 4.3 (11)  & 9.8 (10)\\
  &D@50& 10.31 (55)  & 50.19 (127)  & 44.64 (50)\\
\midrule
\multirow{3}{*}{Mega}
  &D@3& 0.94 (9)    & 1.58 (4) & 3.5 (4) \\
  &D@10& 2.1 (11)   & 0.4 (1) & 9.8 (11) \\
  &D@50& 10.31 (55)    & 0.4 (1) & 73.21 (87) \\
\midrule
\multirow{3}{*}{Llama}
  &D@3& 0.75 (4)  & 1.58 (4)  & 3.5 (4) \\
  &D@10& 2.1 (11)  & 4.3 (11) & 9.8 (11)\\
  &D@50& 10.69 (57)  & 66.79 (169) & 88.39 (99) \\
\midrule
\multirow{3}{*}{Mistral}
  &D@3& 0.75 (4) & 1.58 (4) & 3.5 (4)\\
  &D@10& 2.1 (11)  & 4.3 (11)  & 9.8 (11)\\
  &D@50& 11.06 (59) & 67.19 (170) & 89.28 (92)\\
 \bottomrule
\end{tabular}}
\caption{Higher values represent more diverse retrievals. The numbers in parentheses represent the number of unique emotional events retrieved after deduplication.}
\label{tab:diversity_results}
\end{table}

\subsection{Diversity of Retrieved Emotion Events after De-duplication}
We compare the capacity of LLMs on retrieving diverse events for given emotion, as shown in Table \ref{tab:diversity_results}. For each model, we report the results on D@3, D@10 and D@50 for Joy, Sad and Angry. We can observe that most models perform not well on diverse emotional event retrieval. Mistral and Mega show the best overall performance at high diversity levels, particularly excelling in retrieving unique angry events. BERT performs well for Anger at D@50 but lags behind in Joy and Sadness. Models like T5 and GPT demonstrate weaker performance in capturing the diversity of emotions, especially for Sadness.

\section{Conclusion}
In this study, we assessed the performance of various Large Language Models (LLMs) in retrieving emotional events across three categories—Joy, Sadness, and Anger—at different precision levels (P@3, P@10, and P@50). The results reveal that models demonstrate varying strengths and weaknesses depending on the emotion type and dataset size. These performance disparities across models and emotional categories underscore the need for further advancements in emotion recognition, particularly for underrepresented emotions such as Sadness. Future research should prioritize the development of more balanced models capable of accurately handling a wider range of emotional content with consistently high precision across all categories.

\section*{Limitation}
The experiments are conducted on a specific dataset or a limited set of emotional scenarios, which may constrain the generalizability of the findings. Model performance could vary substantially when applied to more diverse datasets or tested with different types of emotional expressions, such as multimodal inputs including text, audio, or video.

\bibliography{acl_latex}

\end{document}